# Combining Trigram-based and Feature-based Methods for Context-Sensitive Spelling Correction

**Andrew R. Golding** and **Yves Schabes**
Mitsubishi Electric Research Laboratories
201 Broadway
Cambridge, MA 02139
{golding, schabes}@merl.com

## Abstract

This paper addresses the problem of correcting spelling errors that result in valid, though unintended words (such as *peace* and *piece*, or *quiet* and *quite*) and also the problem of correcting particular word usage errors (such as *amount* and *number*, or *among* and *between*). Such corrections require contextual information and are not handled by conventional spelling programs such as Unix *spell*. First, we introduce a method called *Trigrams* that uses part-of-speech trigrams to encode the context. This method uses a small number of parameters compared to previous methods based on word trigrams. However, it is effectively unable to distinguish among words that have the same part of speech. For this case, an alternative feature-based method called *Bayes* performs better; but Bayes is less effective than Trigrams when the distinction among words depends on syntactic constraints. A hybrid method called *Tribayes* is then introduced that combines the best of the previous two methods. The improvement in performance of Tribayes over its components is verified experimentally. Tribayes is also compared with the grammar checker in Microsoft Word, and is found to have substantially higher performance.

## 1 Introduction

Spelling correction has become a very common technology and is often not perceived as a problem where progress can be made. However, conventional spelling checkers, such as Unix *spell*, are concerned only with spelling errors that result in words that cannot be found in a word list of a given language. One analysis has shown that up to 15% of spelling errors that result from elementary typographical errors (character insertion, deletion, or transposition) yield another valid word in the language (Peterson, 1986). These errors remain undetected by traditional spelling checkers. In addition to typographical errors, words that can be easily confused with each other (for instance, the homophones *peace* and *piece*) also remain undetected. Recent studies of actual observed spelling errors have estimated that overall, errors resulting in valid words account for anywhere from 25% to over 50% of the errors, depending on the application (Kukich, 1992).

We will use the term *context-sensitive spelling correction* to refer to the task of fixing spelling errors that result in valid words, such as:

(1) ∗ Can I have a *peace* of cake?

where *peace* was typed when *piece* was intended. The task will be cast as one of lexical disambiguation: we are given a predefined collection of *confusion sets*, such as {*peace, piece*}, {*than, then*}, etc., which circumscribe the space of spelling errors to look for. A confusion set means that each word in the set could mistakenly be typed when another word in the set was intended. The task is to predict, given an occurrence of a word in one of the confusion sets, which word in the set was actually intended.

Previous work on context-sensitive spelling correction and related lexical disambiguation tasks has its limitations. Word-trigram methods (Mays, Damerau, and Mercer, 1991) require an extremely large body of text to train the word-trigram model; even with extensive training sets, the problem of sparse data is often acute. In addition, huge word-trigram tables need to be available at run time. Moreover, word trigrams are ineffective at capturing long-distance properties such as discourse topic and tense.

Feature-based approaches, such as Bayesian classifiers (Gale, Church, and Yarowsky, 1993), decision lists (Yarowsky, 1994), and Bayesian hybrids (Golding, 1995), have had varying degrees of success for the problem of context-sensitive spelling correction. However, we report experiments that show that these methods are of limited effectiveness for cases such as {*their, there, they're*} and {*than, then*}, where the predominant distinction to be made among the words is syntactic.

| Confusion set | Train | Test | Most freq. | Base |
|---|---|---|---|---|
| their, there, they're | 3265 | 850 | their | 56.8 |
| than, then | 2096 | 514 | than | 63.4 |
| its, it's | 1364 | 366 | its | 91.3 |
| your, you're | 750 | 187 | your | 89.3 |
| begin, being | 559 | 146 | being | 93.2 |
| passed, past | 307 | 74 | past | 68.9 |
| quiet, quite | 264 | 66 | quite | 83.3 |
| weather, whether | 239 | 61 | whether | 86.9 |
| accept, except | 173 | 50 | except | 70.0 |
| lead, led | 173 | 49 | led | 46.9 |
| cite, sight, site | 115 | 34 | sight | 64.7 |
| principal, principle | 147 | 34 | principle | 58.8 |
| raise, rise | 98 | 39 | rise | 64.1 |
| affect, effect | 178 | 49 | effect | 91.8 |
| peace, piece | 203 | 50 | peace | 44.0 |
| country, county | 268 | 62 | country | 91.9 |
| amount, number | 460 | 123 | number | 71.5 |
| among, between | 764 | 186 | between | 71.5 |

Table 1: Performance of the baseline method for 18 confusion sets. "Train" and "Test" give the number of occurrences of any word in the confusion set in the training and test corpora. "Most freq." is the word in the confusion set that occurred most often in the training corpus. "Base" is the percentage of correct predictions of the baseline system on the test corpus.

In this paper, we first introduce a method called *Trigrams* that uses part-of-speech trigrams to encode the context. This method greatly reduces the number of parameters compared to known methods, which are based on *word* trigrams. This method also has the advantage that training can be done once and for all, and quite manageably, for all confusion sets; new confusion sets can be added later without any additional training. This feature makes Trigrams a very easily expandable system.

Empirical evaluation of the trigram method demonstrates that it performs well when the words to be discriminated have different parts of speech, but poorly when they have the same part of speech. In the latter case, it is reduced to simply guessing whichever word in the confusion set is the most common representative of its part-of-speech class.

We consider an alternative method, *Bayes*, a Bayesian hybrid method (Golding, 1995), for the case where the words have the same part of speech. We confirm experimentally that Bayes and Trigrams have complementary performance, Trigrams being better when the words in the confusion set have different parts of speech, and Bayes being better when they have the same part of speech. We introduce a hybrid method, *Tribayes*, that exploits this complementarity by invoking each method when it is strongest. Tribayes achieves the best accuracy of the methods under consideration in all situations.

To evaluate the performance of Tribayes with respect to an external standard, we compare it to the grammar checker in Microsoft Word. Tribayes is found to have substantially higher performance.

This paper is organized as follows: first we present the methodology used in the experiments. We then discuss the methods mentioned above, interleaved with experimental results. The comparison with Microsoft Word is then presented. The final section concludes.

## 2 Methodology

Each method will be described in terms of its operation on a single confusion set $C = \{w_1, \ldots, w_n\}$; that is, we will say how the method disambiguates occurrences of words $w_1$ through $w_n$. The methods handle multiple confusion sets by applying the same technique to each confusion set independently.

Each method involves a training phase and a test phase. We trained each method on 80% (randomly selected) of the Brown corpus (Kučera and Francis, 1967) and tested it on the remaining 20%. All methods were run on a collection of 18 confusion sets, which were largely taken from the list of "Words Commonly Confused" in the back of Random House (Flexner, 1983). The confusion sets were selected on the basis of being frequently-occurring in Brown, and representing a variety of types of errors, including homophone confusions (e.g., {*peace*, *piece*}) and grammatical mistakes (e.g., {*among*, *between*}). A few confusion sets not in Random House were added, representing typographical errors (e.g., {*begin*, *being*}). The confusion sets appear in Table 1.

## 3 Baseline

As an indicator of the difficulty of the task, we compared each of the methods to the method which ignores the context in which the word occurred, and just guesses based on the priors.

Table 1 shows the performance of the baseline method for the 18 confusion sets.

## 4 Trigrams

Mays, Damerau, and Mercer (1991) proposed a word-trigram method for context-sensitive spelling correction based on the noisy channel model. Since this method is based on word trigrams, it requires an enormous training corpus to fit all of these parameters accurately; in addition, at run time it requires extensive system resources to store and manipulate the resulting huge word-trigram table.

In contrast, the method proposed here uses part-of-speech trigrams. Given a target occurrence of a word to correct, it substitutes in turn each word in the confusion set into the sentence. For each substitution, it calculates the probability of the resulting sentence. It selects as its answer the word that gives the highest probability.

More precisely, assume that the word $w_k$ occurs in a sentence $W = w_1 \ldots w_k \ldots w_n$, and that $w'_k$ is a word we are considering substituting for it, yielding sentence $W'$. Word $w'_k$ is then preferred over $w_k$ iff $P(W') > P(W)$, where $P(W)$ and $P(W')$ are the probabilities of sentences $W$ and $W'$ respectively.[1] We calculate $P(W)$ using the tag sequence of $W$ as an intermediate quantity, and summing, over all possible tag sequences, the probability of the sentence with that tagging; that is:

$$P(W) = \sum_T P(W, T)$$

where $T$ is a tag sequence for sentence $W$.

The above probabilities are estimated as is traditionally done in trigram-based part-of-speech tagging (Church, 1988; DeRose, 1988):

$$\begin{align} P(W, T) &= P(W|T)P(T) \quad (1) \\ &= \prod_i P(w_i|t_i) \prod_i P(t_i|t_{i-2}t_{i-1}) \quad (2) \end{align}$$

where $T = t_1 \ldots t_n$, and $P(t_i|t_{i-2}t_{i-1})$ is the probability of seeing a part-of-speech tag $t_i$ given the two preceding part-of-speech tags $t_{i-2}$ and $t_{i-1}$. Equations 1 and 2 will also be used to tag sentences $W$ and $W'$ with their most likely part-of-speech sequences. This will allow us to determine the tag that would be assigned to each word in the confusion set when substituted into the target sentence.

Table 2 gives the results of the trigram method (as well as the Bayesian method of the next section) for the 18 confusion sets.[2] The results are broken down into two cases: "Different tags" and "Same tags". A target occurrence is put in the latter iff all words in the confusion set would have the same tag when substituted into the target sentence. In the "Different tags" condition, Trigrams generally does well, outscoring Bayes for all but 3 confusion sets — and in each of these cases, making no more than 3 errors more than Bayes.

In the "Same tags" condition, however, Trigrams performs only as well as Baseline. This follows from Equations 1 and 2: when comparing $P(W)$ and $P(W')$, the dominant term corresponds to the most likely tagging; and in this term, if the target word $w_k$ and its substitute $w'_k$ have the same tag $t$, then the comparison amounts to comparing $P(w_k|t)$ and $P(w'_k|t)$. In other words, the decision reduces to which of the two words, $w_k$ and $w'_k$, is the more common representative of part-of-speech class $t$.[3]

## 5 Bayes

The previous section showed that the part-of-speech trigram method works well when the words in the confusion set have different parts of speech, but essentially cannot distinguish among the words if they have the same part of speech. In this case, a more effective approach is to learn *features* that characterize the different contexts in which each word tends to occur. A number of feature-based methods have been proposed, including Bayesian classifiers (Gale, Church, and Yarowsky, 1993), decision lists (Yarowsky, 1994), Bayesian hybrids (Golding, 1995), and, more recently, a method based on the Winnow multiplicative weight-updating algorithm (Golding and Roth, 1996). We adopt the Bayesian hybrid method, which we will call *Bayes*, having experimented with each of the methods and found Bayes to be among the best-performing for the task at hand. This method has been described elsewhere (Golding, 1995) and so will only be briefly reviewed here; however, the version used here uses an improved smoothing technique, which is mentioned briefly below.

---

[1] To enable fair comparisons between sequences of different length (as when considering *maybe* and *may be*), we actually compare the per-word geometric mean of the sentence probabilities. Otherwise, the shorter sequence will usually be preferred, as shorter sequences tend to have higher probabilities than longer ones.

[2] In the experiments reported here, the trigram method was run using the tag inventory derived from the Brown corpus, except that a handful of common function words were tagged as themselves, namely: *except, than, then, to, too,* and *whether*.

[3] In a few cases, however, Trigrams does not get exactly the same score as Baseline. This can happen when the words in the confusion set have more than one tag in common; e.g., for {*affect, effect*}, the words can both be nouns or verbs. Trigrams may then choose differently when the words are tagged as nouns versus verbs, whereas Baseline makes the same choice in all cases.

| Confusion set | Different tags | | | | Same tags | | | |
|---|---|---|---|---|---|---|---|---|
| | Break-down | System scores | | | Break-down | System scores | | |
| | | Base | T | B | | Base | T | B |
| their, there, they're | 100 | 56.8 | 97.6 | 94.4 | 0 | — | — | — |
| than, then | 100 | 63.4 | 94.9 | 93.2 | 0 | — | — | — |
| its, it's | 100 | 91.3 | 98.1 | 95.9 | 0 | — | — | — |
| your, you're | 100 | 89.3 | 98.9 | 89.8 | 0 | — | — | — |
| begin, being | 100 | 93.2 | 97.3 | 91.8 | 0 | — | — | — |
| passed, past | 100 | 68.9 | 95.9 | 89.2 | 0 | — | — | — |
| quiet, quite | 100 | 83.3 | 95.5 | 89.4 | 0 | — | — | — |
| weather, whether | 100 | 86.9 | 93.4 | 96.7 | 0 | — | — | — |
| accept, except | 100 | 70.0 | 82.0 | 88.0 | 0 | — | — | — |
| lead, led | 100 | 46.9 | 83.7 | 79.6 | 0 | — | — | — |
| cite, sight, site | 100 | 64.7 | 70.6 | 73.5 | 0 | — | — | — |
| principal, principle | 29 | 0.0 | 100.0 | 70.0 | 71 | 83.3 | 83.3 | 91.7 |
| raise, rise | 8 | 100.0 | 100.0 | 100.0 | 92 | 61.1 | 61.1 | 72.2 |
| affect, effect | 6 | 100.0 | 100.0 | 66.7 | 94 | 91.3 | 93.5 | 97.8 |
| peace, piece | 2 | 0.0 | 100.0 | 100.0 | 98 | 44.9 | 42.9 | 89.8 |
| country, county | 0 | — | — | — | 100 | 91.9 | 91.9 | 85.5 |
| amount, number | 0 | — | — | — | 100 | 71.5 | 73.2 | 82.9 |
| among, between | 0 | — | — | — | 100 | 71.5 | 71.5 | 75.3 |

Table 2: Performance of the component methods, Baseline (Base), Trigrams (T), and Bayes (B). System scores are given as percentages of correct predictions. The results are broken down by whether or not all words in the confusion set would have the same tagging when substituted into the target sentence. The "Breakdown" columns show the percentage of examples that fall under each condition.

Bayes uses two types of features: context words and collocations. Context-word features test for the presence of a particular word within $\pm k$ words of the target word; collocations test for a pattern of up to $\ell$ contiguous words and/or part-of-speech tags around the target word. Examples for the confusion set {*dairy, diary*} include:

(2) *milk* within $\pm 10$ words
(3) *in* POSS-DET __

where (2) is a context-word feature that tends to imply *dairy*, while (3) is a collocation implying *diary*. Feature (3) includes the tag POSS-DET for possessive determiners (*his, her*, etc.), and matches, for example, the sequence *in his*[4] in:

(4) He made an entry in his *diary*.

Bayes learns these features from a training corpus of correct text. Each time a word in the confusion set occurs in the corpus, Bayes proposes every feature that matches the context — one context-word feature for every distinct word within $\pm k$ words of the target word, and one collocation for every way of expressing a pattern of up to $\ell$ contiguous elements. After working through the whole training corpus, Bayes collects and returns the set of features proposed. Pruning criteria may be applied at this point to eliminate features that are based on insufficient data, or that are ineffective at discriminating among the words in the confusion set.

At run time, Bayes uses the features learned during training to correct the spelling of target words. Let $\mathcal{F}$ be the set of features that match a particular target occurrence. Suppose for a moment that we were applying a *naive* Bayesian approach. We would then calculate the probability that each word $w_i$ in the confusion set is the correct identity of the target word, given that we have observed features $\mathcal{F}$, using Bayes' rule with the independence assumption:

$$P(w_i|\mathcal{F}) = \left(\prod_{f \in \mathcal{F}} P(f|w_i)\right) \frac{P(w_i)}{P(\mathcal{F})}$$

where each probability on the right-hand side is calculated by a maximum-likelihood estimate (MLE) over the training set. We would then pick as our answer the $w_i$ with the highest $P(w_i|\mathcal{F})$. The method presented here differs from the naive approach in two respects: first, it does not assume independence among features, but rather has heuristics for detecting strong dependencies, and resolving them by deleting features until it is left with a reduced set $\mathcal{F}'$ of (relatively) independent features, which are then used in place of $\mathcal{F}$ in the formula above. Second, to estimate the $P(f|w_i)$ terms, rather than using a simple MLE, it performs *smoothing* by interpolating between the MLE of $P(f|w_i)$ and the MLE of the unigram probability, $P(f)$. These enhancements greatly improve the performance of Bayes over the naive Bayesian approach.

---

[4] A tag is taken to match a word in the sentence iff the tag is a member of the word's set of possible part-of-speech tags. Tag sets are used, rather than actual tags, because it is in general impossible to tag the sentence uniquely at spelling-correction time, as the identity of the target word has not yet been established.

The results of Bayes are shown in Table 2.[5] Generally speaking, Bayes does worse than Trigrams when the words in the confusion set have different parts of speech. The reason is that, in such cases, the predominant distinction to be made among the words is *syntactic*; and the trigram method, which brings to bear part-of-speech knowledge for the whole sentence, is better equipped to make this distinction than Bayes, which only tests up to two syntactic elements in its collocations. Moreover, Bayes' use of context-word features is arguably misguided here, as context words pick up differences in topic and tense, which are irrelevant here, and in fact tend to degrade performance by detecting spurious differences. In a few cases, such as {*begin*, *being*}, this effect is enough to drive Bayes slightly below Baseline.[6]

For the condition where the words have the same part of speech, Table 2 shows that Bayes almost always does better than Trigrams. This is because, as discussed above, Trigrams is essentially acting like Baseline in this condition. Bayes, on the other hand, learns features that allow it to discriminate among the particular words at issue, regardless of their part of speech. The one exception is {*country*, *county*}, for which Bayes scores somewhat below Baseline. This is another case in which context words actually hurt Bayes, as running it without context words again improved its performance to the Baseline level.

## 6 Tribayes

The previous sections demonstrated the complementarity between Trigrams and Bayes: Trigrams works best when the words in the confusion set do not all have the same part of speech, while Bayes works best when they do. This complementarity leads directly to a hybrid method, *Tribayes*, that gets the best of each. It applies Trigrams first; in the process, it ascertains whether all the words in the confusion set would have the same tag when substituted into the target sentence. If they do not, it accepts the answer provided by Trigrams; if they do, it applies Bayes.

Two points about the application of Bayes in the hybrid method: first, Bayes is now being asked to distinguish among words only when they have the same part of speech. It should be trained accordingly — that is, only on examples where the words have the same part of speech. The Bayes component of the hybrid will therefore be trained on a subset of the examples that would be used for training the stand-alone version of Bayes.

The second point about Bayes is that, like Trigrams, it sometimes makes uninformed decisions — decisions based only on the priors. For Bayes, this happens when none of its features matches the target occurrence. Since, for now, we do not have a good "third-string" algorithm to call when both Trigrams and Bayes fall by the wayside, we content ourselves with the guess made by Bayes in such situations.

Table 3 shows the performance of Tribayes compared to its components. In the "Different tags" condition, Tribayes invokes Trigrams, and thus scores identically. In the "Same tags" condition, Tribayes invokes Bayes. It does not necessarily score the same, however, because, as mentioned above, it is trained on a subset of the examples that stand-alone Bayes is trained on. This can lead to higher or lower performance — higher because the training examples are more homogeneous (representing only cases where the words have the same part of speech); lower because there may not be enough training examples to learn from. Both effects show up in Table 3.

Table 4 summarizes the overall performance of all methods discussed. It can be seen that Trigrams and Bayes each have their strong points. Tribayes, however, achieves the maximum of their scores, by and large, the exceptions being due to cases where one method or the other had an unexpectedly low score (discussed in Sections 4 and 5). The confusion set {*raise*, *rise*} demonstrates (albeit modestly) the ability of the hybrid to outscore *both* of its components, by putting together the performance of the better component for both conditions.

## 7 Comparison with Microsoft Word

The previous section evaluated the performance of Tribayes with respect to its components, and showed that it got the best of both. In this section, we calibrate this overall performance by comparing Tribayes with Microsoft Word (version 7.0), a widely used word-processing system whose grammar checker represents the state of the art in commercial context-sensitive spelling correction.

Unfortunately we cannot evaluate Word using "prediction accuracy" (as we did above), as we do not always have access to the system's predictions — sometimes it *suppresses* its predictions in an effort to filter out the bad ones. Instead, in this section we will use two parameters to evaluate system performance: system accuracy when tested on *correct* usages of words, and system accuracy on *incorrect* usages. Together, these two parameters give a complete picture of system performance: the score on correct usages measures the system's rate of false negative errors (changing a right word to a wrong one), while the score on incorrect usages measures false positives (failing to change a wrong word to a right one). We will not attempt to combine these two parameters into a single measure of system "goodness", as the appropriate combination varies for different users, depending on the user's typing accuracy and tolerance of false negatives and positives.

---

[5] For the experiments reported here, Bayes was configured as follows: $k$ (the half-width of the window of context words) was set to 10; $\ell$ (the maximum length of a collocation) was set to 2; feature strength was measured using the reliability metric; pruning of collocations at training time was enabled; and pruning of context words was minimal — context words were pruned only if they had fewer than 2 occurrences or non-occurrences.

[6] We confirmed this by running Bayes without context words (i.e., with collocations only). Its performance was then always at or above Baseline.

| Confusion set | Different tags | | | Same tags | | |
|---|---|---|---|---|---|---|
| | Break-down | System scores | | Break-down | System scores | |
| | | T | TB | | B | TB |
| their, there, they're | 100 | 97.6 | 97.6 | 0 | — | — |
| than, then | 100 | 94.9 | 94.9 | 0 | — | — |
| its, it's | 100 | 98.1 | 98.1 | 0 | — | — |
| your, you're | 100 | 98.9 | 98.9 | 0 | — | — |
| begin, being | 100 | 97.3 | 97.3 | 0 | — | — |
| passed, past | 100 | 95.9 | 95.9 | 0 | — | — |
| quiet, quite | 100 | 95.5 | 95.5 | 0 | — | — |
| weather, whether | 100 | 93.4 | 93.4 | 0 | — | — |
| accept, except | 100 | 82.0 | 82.0 | 0 | — | — |
| lead, led | 100 | 83.7 | 83.7 | 0 | — | — |
| cite, sight, site | 100 | 70.6 | 70.6 | 0 | — | — |
| principal, principle | 29 | 100.0 | 100.0 | 71 | 91.7 | 83.3 |
| raise, rise | 8 | 100.0 | 100.0 | 92 | 72.2 | 75.0 |
| affect, effect | 6 | 100.0 | 100.0 | 94 | 97.8 | 95.7 |
| peace, piece | 2 | 100.0 | 100.0 | 98 | 89.8 | 89.8 |
| country, county | 0 | — | — | 100 | 85.5 | 85.5 |
| amount, number | 0 | — | — | 100 | 82.9 | 82.9 |
| among, between | 0 | — | — | 100 | 75.3 | 75.3 |

Table 3: Performance of the hybrid method, Tribayes (TB), as compared with Trigrams (T) and Bayes (B). System scores are given as percentages of correct predictions. The results are broken down by whether or not all words in the confusion set would have the same tagging when substituted into the target sentence. The "Breakdown" columns give the percentage of examples under each condition.

| Confusion set | System scores | | | |
|---|---|---|---|---|
| | Base | T | B | TB |
| their, there, they're | 56.8 | 97.6 | 94.4 | 97.6 |
| than, then | 63.4 | 94.9 | 93.2 | 94.9 |
| its, it's | 91.3 | 98.1 | 95.9 | 98.1 |
| your, you're | 89.3 | 98.9 | 89.8 | 98.9 |
| begin, being | 93.2 | 97.3 | 91.8 | 97.3 |
| passed, past | 68.9 | 95.9 | 89.2 | 95.9 |
| quiet, quite | 83.3 | 95.5 | 89.4 | 95.5 |
| weather, whether | 86.9 | 93.4 | 96.7 | 93.4 |
| accept, except | 70.0 | 82.0 | 88.0 | 82.0 |
| lead, led | 46.9 | 83.7 | 79.6 | 83.7 |
| cite, sight, site | 64.7 | 70.6 | 73.5 | 70.6 |
| principal, principle | 58.8 | 88.2 | 85.3 | 88.2 |
| raise, rise | 64.1 | 64.1 | 74.4 | 76.9 |
| affect, effect | 91.8 | 93.9 | 95.9 | 95.9 |
| peace, piece | 44.0 | 44.0 | 90.0 | 90.0 |
| country, county | 91.9 | 91.9 | 85.5 | 85.5 |
| amount, number | 71.5 | 73.2 | 82.9 | 82.9 |
| among, between | 71.5 | 71.5 | 75.3 | 75.3 |

Table 4: Overall performance of all methods: Baseline (Base), Trigrams (T), Bayes (B), and Tribayes (TB). System scores are given as percentages of correct predictions.

The test sets for the correct condition are the same ones used earlier, based on 20% of the Brown corpus. The test sets for the incorrect condition were generated by corrupting the correct test sets; in particular, each correct occurrence of a word in the confusion set was replaced, in turn, with each other word in the confusion set, yielding $n-1$ incorrect occurrences for each correct occurrence (where $n$ is the size of the confusion set). We will also refer to the incorrect condition as the *corrupted* condition.

To run Microsoft Word on a particular test set, we started by disabling error checking for all error types except those needed for the confusion set at issue. This was done to avoid confounding effects. For $\{their, there, they're\}$, for instance, we enabled "word usage" errors (which include substitutions of *their* for *there*, etc.), but we disabled "contractions" (which include replacing *they're* with *they are*). We then invoked the grammar checker, accepting every suggestion offered. Sometimes errors were pointed out but no correction given; in such cases, we skipped over the error. Sometimes the suggestions led to an infinite loop, as with the sentence:

(5) Be sure it's out when you leave.

where the system alternately suggested replacing *it's* with *its* and vice versa. In such cases, we accepted the first suggestion, and then moved on.

Unlike Word, Tribayes, as presented above, is purely a predictive system, and never suppresses its suggestions. This is somewhat of a handicap in the comparison, as Word can achieve higher scores in the correct condition by suppressing its weaker suggestions (albeit at the cost of lowering its scores in the corrupted condition). To put Tribayes on an equal footing, we added a postprocessing step in which it uses *thresholds* to decide whether to suppress its suggestions. A suggestion is allowed to go through iff the ratio of the probability of the word being suggested to the probability of the word that appeared originally in the sentence is above a threshold. The probability associated with each word is the per-word sentence probability in the case of Trigrams, or the conditional probability $P(w_i|\mathcal{F})$ in the case of Bayes. The thresholds are set in a preprocessing phase based on the training set (80% of Brown, in our case). A single tunable parameter controls how steeply the thresholds are set; for the study here, this parameter was set to the middle of its useful range, providing a fairly neutral balance between reducing false negatives and increasing false positives.

The results of Word and Tribayes for the 18 confusion sets appear in Table 5. Six of the confusion sets (marked with asterisks in the table) are not handled by Word; Word's scores in these cases are 100% for the correct condition and 0% for the corrupted condition, which are the scores one gets by never making a suggestion. The opposite behavior — *always* suggesting a different word — would result in scores of 0% and 100% (for a confusion set of size 2). Although this behavior is never observed in its extreme form, it is a good approximation of Word's behavior in a few cases, such as $\{principal, principle\}$, where it scores 12% and 94%. In general, Word achieves a high score in either the correct or the corrupted condition, but not both at once.

Tribayes compares quite favorably with Word in this experiment. In both the correct and corrupted conditions, Tribayes' scores are mostly higher (often by a wide margin) or the same as Word's; in the cases where they are lower in one condition, they are almost always considerably higher in the other. The one exception is $\{raise, rise\}$, where Tribayes and Word score about the same in both conditions.

## 8 Conclusion

Spelling errors that result in valid, though unintended words, have been found to be very common in the production of text. Such errors were thought to be too difficult to handle and remain undetected in conventional spelling checkers. This paper introduced Trigrams, a part-of-speech trigram-based method, that improved on previous trigram methods, which were word-based, by greatly reducing the number of parameters. The method was supplemented by Bayes, a method that uses context features to discriminate among the words in the confusion set. Trigrams and Bayes were shown to have complementary strengths. A hybrid method, Tribayes, was then introduced to exploit this complementarity by applying Trigrams when the words in the confusion set do not have the same part of speech, and Bayes when they do. Tribayes thereby gets the best of both methods, as was confirmed experimentally. Tribayes was also compared with the grammar checker in Microsoft Word, and was found to have substantially higher performance.

Tribayes is being used as part of a grammar-checking system we are currently developing. We are presently working on elaborating the system's threshold model; scaling up the number of confusion sets that can be handled efficiently; and acquiring confusion sets (or confusion matrices) automatically.

| Confusion set | Tribayes | | Microsoft Word | |
|---|---|---|---|---|
| | Correct | Corrupted | Correct | Corrupted |
| their, there, they're | 99.4 | 87.6 | 98.8 | 59.8 |
| than, then | 97.9 | 85.8 | 100.0 | 22.2 |
| its, it's | 99.5 | 92.1 | 96.2 | 73.0 |
| your, you're | 98.9 | 98.4 | 98.9 | 79.1 |
| begin, being | 100.0 | 84.2 | 100.0 * | 0.0 * |
| passed, past | 100.0 | 92.4 | 37.8 | 86.5 |
| quiet, quite | 100.0 | 72.7 | 100.0 * | 0.0 * |
| weather, whether | 100.0 | 65.6 | 100.0 * | 0.0 * |
| accept, except | 90.0 | 70.0 | 74.0 | 36.0 |
| lead, led | 87.8 | 81.6 | 100.0 * | 0.0 * |
| cite, sight, site | 100.0 | 35.3 | 17.6 | 66.2 |
| principal, principle | 94.1 | 73.5 | 11.8 | 94.1 |
| raise, rise | 92.3 | 48.7 | 92.3 | 51.3 |
| affect, effect | 98.0 | 93.9 | 100.0 | 77.6 |
| peace, piece | 96.0 | 74.0 | 36.0 | 88.0 |
| country, county | 90.3 | 80.6 | 100.0 * | 0.0 * |
| amount, number | 91.9 | 68.3 | 100.0 * | 0.0 * |
| among, between | 88.7 | 54.8 | 97.8 | 0.0 |

Table 5: Comparison of Tribayes with Microsoft Word. System scores are given for two test sets, one containing correct usages, and the other containing incorrect (corrupted) usages. Scores are given as percentages of correct answers. Asterisks mark confusion sets that are not handled by Microsoft Word.